\documentclass{bmvc2k}
\usepackage[ruled,linesnumbered]{algorithm2e}
\usepackage{amsmath,amsfonts,stmaryrd,amssymb,bm,bbm}
\usepackage{pifont}
\newcommand{\cmark}{\ding{51}}
\newcommand{\xmark}{\ding{55}}
\DeclareMathOperator*{\argmax}{arg\,max}

\title{Dual Moving Average Pseudo-Labeling for Source-Free Inductive Domain Adaptation}

\addauthor{Hao Yan}{haoyan6@cmail.carleton.ca}{}
\addauthor{Yuhong Guo}{yuhong.guo@carleton.ca}{}

\addinstitution{
 Carleton University\\
 Ottawa, Canada
}
\addinstitution{
  CIFAR AI Chair\\
  Amii, Canada
}

\runninghead{Yan, Guo}{Source-Free Inductive Domain Adaptation}

\begin{document}

\maketitle

\begin{abstract}
Unsupervised domain adaptation reduces the reliance on data annotation 
	in deep learning by adapting 
	knowledge from a source to a target domain. For privacy and efficiency concerns, source-free domain adaptation 
	extends unsupervised domain adaptation by adapting
	a pre-trained source model to an 
	unlabeled target domain without accessing the source data. 
	However, most existing source-free domain adaptation methods to date
	focus on the transductive setting, where the target training set is also the testing set. In this paper, we address source-free domain adaptation 
	in the more realistic inductive setting, where the target training and testing sets are mutually exclusive. We propose a new semi-supervised fine-tuning 
	method named Dual Moving Average Pseudo-Labeling (DMAPL) for source-free inductive domain adaptation. 
	We first split the unlabeled training set in the target domain
	into a pseudo-labeled confident subset and an unlabeled less-confident subset according to the prediction confidence scores from the pre-trained source model. 
	Then we propose a soft-label moving-average updating strategy for the unlabeled subset based on a moving-average prototypical classifier, 
	which gradually adapts the source model towards the target domain. 
	Experiments show that our proposed method achieves state-of-the-art performance and outperforms previous methods by large margins.
\end{abstract}

\section{Introduction}
\label{sec:intro}
Training deep models
requires large scale datasets with accurate annotations. 
Data annotation however can be difficult
in many real-world domains. 
Unsupervised domain adaptation (UDA) aims at reducing the dependence on data annotation in a target domain
with the help of another already labeled source dataset. 
Existing UDA methods have attempted to learn a good prediction model for the target domain
from both the labeled data in the source domain and the unlabeled data in the target domain 
by bridging the domain divergence gap through 
adversarial learning \cite{ganin2015unsupervised, mingsheng2018}, 
metric minimization
\cite{tzeng2014deep, lee2019sliced},
or regularized semi-supervised learning
\cite{jin2020minimum, rangwani2022closer, berthelot2022adamatch}. 
However, in many domains that involve personal or commercial data, 
organizations usually prefer to release a pre-trained model instead of
sharing a labeled source dataset due to data privacy concern. 
Moreover, storing and transferring large scale source dataset are way less efficient than working with a pre-trained source model. 
For such privacy and efficiency reasons, source-free domain adaptation (SFDA),
which aims at learning a target model based on the pre-trained source model and the unlabeled target data,
has recently become an emerging topic. 
Most existing SFDA methods
have adopted
a semi-supervised fine-tuning framework, 
where unlabeled target data are used 
to fine-tune the pre-trained source model. 
They exploit the unlabeled data in the target domain 
by either assigning pseudo-labels 
to the unlabeled target data 
\cite{liang2020we, yang2021exploiting},
or utilizing regularization terms
such as mutual information maximization \cite{liang2020we} and contrastive learning \cite{huang2021model}
to learn from the unlabeled target distribution. 
Some other works have also attempted to
generate labeled target data or 
surrogate source data
from the pre-trained source model 
by using generative models \cite{li2020model} or optimization strategies \cite{yan2021source}. 
Nevertheless, these existing
SFDA methods have all focused on a 
transductive setting, 
where the unlabeled training set in the target domain is also used as the testing set,
and hence a method that overfits the unlabeled target training set can demonstrate good performance
but has poor generalizability.

In this paper, we consider a more realistic source-free {\em inductive} domain adaptation setting,
where the unlabeled target training set and the testing set are mutually exclusive. 
This setting aims to evaluate SFDA methods in terms of their generalization ability on unseen test data,
which is important for real-world system deployment. 
To tackle this inductive SFDA problem, 
we propose a new semi-supervised fine-tuning 
method named Dual Moving Average Pseudo-Labeling (DMAPL). 
In the proposed method, to prevent over adaptation to the unlabeled target training data, 
we split the unlabeled training set in the target domain into 
two subsets based on the predictions made by the pre-trained source model: 
a confident subset of instances with high confident label prediction scores,
and a less confident subset with low confident label prediction scores. 
As the high confident subset is better aligned with the source model,
we use this subset with the predicted pseudo-labels
as a fixed labeled subset to preserve 
the generalizable prediction properties of the pre-trained source model 
and alleviate the potential overfitting to the unlabeled target training data.  
The less confident subset is then used as an unlabeled subset for semi-supervised model adaptation. 
We design
a soft-label updating strategy to gradually update the soft-labels for the unlabeled subset
in a moving average manner using
a moving-average based prototypical classifier. 
The soft-labels are 
used as the 
pseudo-labels
of the unlabeled subset to fine-tune the prediction model. 

To evaluate the proposed method, we conduct source-free inductive domain adaptation experiments on 
two large domain adaptation benchmark datasets.
Our proposed method achieves state-of-the-art performance and outperforms the existing SFDA methods 
by large margins. 
The experiments also confirm that the relative performance of some existing UDA and SFDA methods
is very different in this inductive domain adaptation setting from the previous reported 
performance in the transductive domain adaptation setting. 
This may inspire future works to investigate domain adaptation in the more realistic inductive setting.

\section{Related Works}
\label{sec:related}

\paragraph{Unsupervised Domain Adaptation.} 
Unsupervised domain adaptation (UDA) aims at learning a target model given labeled source data and unlabeled target data under cross-domain shift in data distribution. Existing UDA methods can be roughly divided into two categories, alignment-based and regularization-based methods. Alignment-based methods try to reduce domain discrepancy at different levels with various alignment techniques based on the theoretical analysis in \cite{ben2010theory}. These alignment techniques include maximum mean discrepancy (MMD) \cite{tzeng2014deep}, adversarial alignment \cite{ganin2015unsupervised}, moment matching \cite{chen2020homm}, 
and optimal transport \cite{lee2019sliced}. 
Alignment can 
be performed at feature-level \cite{mingsheng2018}, input-level \cite{hoffman2018cycada} and output-level \cite{tsai2018learning}. 
Regularization-based approaches usually treat the UDA problem as a semi-supervised learning problem,
while adopting
pseudo-labeling \cite{lee2013pseudo} 
to assign pseudo-labels to unlabeled target data 
for self-training \cite{zou2019confidence, liu2021cycle, berthelot2022adamatch}. 
Prediction smoothness losses based on instances and model-weight perturbations \cite{shu2018a, rangwani2022closer} 
have been used as model regularization terms for unsupervised domain adaptation. 
Some other regularization terms, 
such as entropy minimization \cite{shu2018a, vu2019advent}, nuclear-norm maximization \cite{cui2020towards}, and prediction consistency loss \cite{french2018self},
have also been
used to boost target model performance.

\paragraph{Source-Free Domain Adaptation.} 
Source-free domain adaptation (SFDA) aims at learning a prediction model in the target domain given the pre-trained source model and the unlabeled data from the target domain. Existing SFDA methods are mostly based on semi-supervised fine-tuning,
which uses different tricks 
to fine-tune the pre-trained source model with unlabeled target domain data. 
These methods can be roughly categorized 
into three groups: self-training based methods, data generation based methods, and regularization based methods.
Self-training based methods assign pseudo-labels to 
the target domain data via various techniques including feature clustering \cite{liang2020we, yang2021exploiting, qiu2021source}, naive pseudo-labeling with instance weighting \cite{kundu2020universal, lee2022confidence}, self-labeling \cite{yan2021augmented}, and learning with noisy label \cite{zhang2021unsupervised}. 
Data generation based 
methods generate either labeled data for the
source distribution \cite{hou2021visualizing, yan2021source} or labeled target domain data \cite{li2020model, kurmi2021domain} 
to facilitate domain adaptation.
Regularization-based methods utilize regularization terms to help target model fine-tuning by exploring intrinsic characteristics in the target distribution. This includes mutual information maximization \cite{liang2020we}, contrastive feature learning \cite{xia2021adaptive, huang2021model}, instance and feature level mix-up \cite{kundu2022balancing}. 
Some methods also adopt
additional model components during source model training, such as weight normalization \cite{liang2020we} and domain attention \cite{yang2021generalized}. 
However, all the existing SFDA methods address the transductive setting,
where the unlabeled training set from the target domain is also the testing set. 
In this setting, a method that overfits the unlabeled target training set can 
demonstrate good performance but 
is not able to generalize well on unseen test data. 
In this paper, we consider an inductive setting, 
where the target training and testing sets are mutually exclusive,
aiming to develop a more realistic experimental setup for SFDA.

\section{Proposed Method}
\label{sec:method}

\begin{figure}[t]
    \centering
    \includegraphics[width=\textwidth]{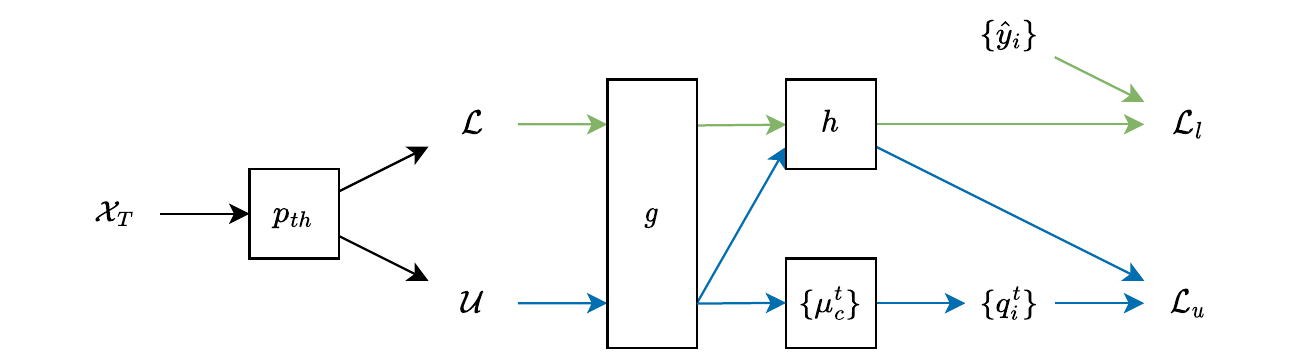}
    \caption{Illustration of the proposed method. 
The target training data are first split into a pseudo-labeled confident subset ($\mathcal{L}$) 
	and a less-confident unlabeled subset ($\mathcal{U}$) 
	based on the pre-trained source model.
The pseudo-labeled confident subset acts as trusty supervision to prevent over adaptation.
	The unlabeled subset is gradually updated to fine-tune the prediction model 
	($f=h\circ g$)
through dual moving average update. 
	}
    \label{fig:diagram}
\end{figure}

This paper addresses 
the source-free inductive domain adaptation problem, 
where a pre-trained source model $f_S$ and an unlabeled target training set $\mathcal{X}_T$ are given. 
The goal is to train a target model $f$ 
that can generalize well on a target testing set that is unseen during training.
To tackle this problem, we propose a new semi-supervised fine-tuning approach 
to adapt the source model with the unlabeled target training data. 
To prevent over adaptation, 
we split the unlabeled target training data into a {\em pseudo-labeled subset} 
with high prediction confidence and an {\em unlabeled subset} with less confident predictions 
based on the pre-trained source model. 
The pseudo-labeled subset acts as trusty supervision for the target model fine-tuning. 
As the pseudo-labels of the unlabeled subset are much more noisy, 
we propose a soft-label moving-average updating method based on a moving-averaged prototypical classifier. 
The proposed method is illustrated in Figure~\ref{fig:diagram} 
and elaborated in the following subsections.

\subsection{Target Training Data Splitting}

Learning with pseudo-labels is a classical semi-supervised learning strategy, 
where the predicted labels by the current model are used as pseudo-labels 
for the unlabeled instances 
for further model update/fine-tuning. 
These pseudo-labels are usually noisy if the current model is not trained well, which will cause error accumulation during the iterative training process. In the source-free inductive domain adaptation setting,
such gradual pseudo-labeling updates could make the model overfit 
the unlabeled target training data and hurt its generalization ability.
On the other hand, pseudo-labeling with threshold only selects instances with highly confident predictions and ignores those with less confident predictions,
which could lead to a loss of valuable data distribution information.

In view of these issues, we propose to split the unlabeled target training data $\mathcal{X}_T$ into 
a confident subset $\mathcal{L}$ 
 and a less-confident subset $\mathcal{U}$ 
based on the pre-trained source model and 
then exploit them in separate ways. 
Specifically, for each unlabeled target training instance ${\bf x}_i\in \mathcal{X}_T$,
we use $p(y|{\bf x}_i;{\bf \theta}_{f_S})$ to denote 
the predicted probability of
$\bf{x}_i$ belonging to the $y$-th class
by the pre-trained source model $f_S$.
By defining a prediction threshold $p_{th}$, we then select the instances 
with maximum prediction probability exceeding the threshold $p_{th}$
to form the confident subset $\mathcal{L}$ and others form the less-confident subset $\mathcal{U}$, i.e.
\begin{equation}
    \mathcal{L} = \left\{{\bf x}_i\in \mathcal{X}_T| \max_y p(y|{\bf x}_i;{\bf \theta}_{f_S}) \geq p_{th}\right\}, 
	\qquad \mathcal{U} = \mathcal{X}_T \backslash \mathcal{L}.
\end{equation}
Hence for each instance in the confident subset $\mathcal{L}$,  
its pseudo-label can be assigned with high probability (larger then $p_{th}$)
based on the source model predictions
as follows:
\begin{equation}
    \hat{y}_i = \argmax_y\, p(y|{\bf x}_i;{\bf \theta}_{f_S}),\quad 
	{\bf x}_i \in \mathcal{L}.
\end{equation}
All the instances in $\mathcal{L}$ and their corresponding pseudo-labels 
can then form a pseudo-labeled subset   
$\{({\bf x}_i, \hat{y}_i)| {\bf x}_i \in \mathcal{L}\}$.
If the prediction threshold $p_{th}$ is high enough, 
the high confident subset $\mathcal{L}$ will be well aligned with the pre-trained source model. 
We hence propose to use this subset $\mathcal{L}$ with the predicted pseudo-labels as a fixed {\em labeled subset} 
and use the less confident subset $\mathcal{U}$ as an {\em unlabeled subset} for semi-supervised model fine-tuning.
The unlabeled subset will be used to fine-tune and adapt the source model to the target domain,
while the labeled subset will be used as trusty supervision
to preserve the generalizable prediction properties of the source model
and prevent over adaptation to the unlabeled target subset during the semi-supervised fine-tuning process.
Specifically, we consider the following supervised cross-entropy loss on the labeled subset $\mathcal{L}$:
\begin{equation}
    \mathcal{L}_{l} = \mathbb{E}_{{\bf x}_i\in\mathcal{L}} \left[- \log (p(\hat{y}_i|{\bf x}_i;{\bf \theta}_f))\right],
\end{equation}
where ${\bf \theta}_f$ denotes the parameters of target model $f$,
which is initialized  
from the source model $f_S$ at the very beginning. 

\subsection{Dual Moving Average based Model Fine-Tuning}

By using $\mathcal{L}$ as a fixed labeled subset, we conduct semi-supervised model fine-tuning
with soft pseudo-labeling updates on the unlabeled subset $\mathcal{U}$.  
Inspired by the partial label learning work \cite{wang2022pico}, 
we introduce a dual moving average based soft-label updating scheme 
to assign soft-labels to the instances in the unlabeled subset $\mathcal{U}$. 
Typical deep prediction
models are composed of a convolutional encoder and a linear or MLP classifier head. 
We hence denote the target model $f$ 
as a composition of a feature encoder $g$ and a classifier $h$, i.e. $f=h\circ g$. 
For each instance ${\bf x}_i$, 
$g({\bf x}_i)$ denotes
its feature vector extracted by the encoder $g$.
Based on the extracted feature vectors, we introduce a prototypical classifier with centroids denoted 
as $\{{\bf \mu}_c\}_{c=1}^C$, where $C$ is the number of classes. 
Each centroid ${\bf \mu}_c$ denotes the prototype corresponding to the $c$-th class and 
it is updated with the feature vectors of the instances belonging to the $c$-th class in every batch. 
To calculate the centroids, we first normalize each feature vector $g({\bf x}_i)$ by its L2-norm, 
resulting in 
a normalized vector
${\bf z}_i=g({\bf x}_i)/\|g({\bf x}_i)\|_2$. 
Considering the current mini-batch with labeled instances ${X}_l$ sampled from $\mathcal{L}$ and 
unlabeled instances ${X}_u$ sampled from $\mathcal{U}$, 
the feature mean of the $c$-th class in the current iteration $t$ can be calculated as,
\begin{equation}
    {\bf v}_c^t = \frac{\sum_{{\bf x}_i\in {({X}_l} \cup {X}_u)} \mathbbm{1}(\bar{y}_i=c) \cdot {\bf z}_i}
    {\sum_{{\bf x}_i\in {({X}_l} \cup {X}_u)} \mathbbm{1}(\bar{y}_i=c)},
\end{equation}
where 
$\mathbbm{1}(\cdot)$ denotes an indicator function; 
and $\bar{y}_i$ denotes the pseudo-label for instance ${\bf x}_i$. 
For instances from the labeled subset, 
we always use the pre-assigned pseudo-labels by the initial source model, 
i.e., $\bar{y}_i=\hat{y}_i$ for ${\bf x}_i\in \mathcal{L}$.
For instances from unlabeled subset, 
we use the pseudo-labels predicted by the current model, i.e., $\bar{y}_i=\argmax_y p(y|{\bf x}_i;{\bf \theta}_f)$for ${\bf x}_i\in \mathcal{U}$. 
We then calculate the centroid ${\bf \mu}_c^t$ for the current iteration $t$ 
as the weighted average of the centroid ${\bf \mu}_c^{t-1}$ 
from the previous iteration
and the feature mean ${\bf v}_c^t$ in he current iteration: 
\begin{equation}
    {\bf \mu}_c^t = \text{Normalize}(\alpha {\bf \mu}_c^{t-1} + (1-\alpha) {\bf v}_c^t),
\end{equation}
where $\text{Normalize}(\cdot)$ 
denotes the L2-normalization function
and $\alpha\in(0,1)$ is a moving average coefficient hyperparameter.
Starting with the initial centroid ${\bf \mu}_c^0={\bf 0}$,
the centroid vector will be updated by gradually discounting the previous vector with a factor $\alpha$ 
and adding the current feature mean vector ${\bf v}_c^t$ with a weight $(1-\alpha)$ in each iteration.

The prototypical classifier assigns a new one-hot label vector ${\bf \tilde{y}}_i^t$ 
to each unlabeled instance ${\bf x}_i\in {X}_u$ by placing 1 in the $j$-th entry of the label vector
such that 
the maximum inner-product value is produced between 
the normalized feature vector ${\bf z}_i$ and the centroid vector ${\bf \mu}_j^t$ 
among all centroids $\{{\bf \mu}_1^t, \cdots, {\bf \mu}_C^t\}$, such that
\begin{equation}
	({\bf \tilde{y}}_i^t)_j = \left\{ \begin{matrix}
		1,& j= \argmax_{c\in\{1,\cdots,C\}} {\bf z}_i^\top {\bf \mu}_c^t, \\
        0,& \text{otherwise}.
    \end{matrix}
    \right.
\end{equation}
where $({\bf \tilde{y}}_i^t)_j$ denotes the $j$-th element of the vector ${\bf \tilde{y}}_i^t$.
Note as both vectors are L2-normalized, the inner-product above actually measures the cosine-similarity between the feature vector and the centroid. 
This newly assigned pseudo-label vector however will not be used directly to train the target model as it is still 
very noisy. 
Instead, we use the pseudo-label vectors produced by the prototypical classifier to update the soft-labels
of the unlabeled subset gradually in a moving average manner. 
We denote the soft-label vector for an unlabeled instance ${\bf x}_i$ in the current iteration $t$ 
as ${\bf q}_i^t$, while  ${\bf q}_i^0={\bf 0}$. 
It is updated as the weighted average of the soft-label vector ${\bf q}_i^{t-1}$ from
the previous iteration and the one-hot pseudo-label vector ${\bf \tilde{y}}_i^t$ from the current prototypical classifier: 
\begin{align}
    {\bf q}_i^t = \beta {\bf q}_i^{t-1} + (1-\beta) {\bf \tilde{y}}_i^t,
	\label{qupdate}
\end{align}
where $\beta\in (0,1)$ is another coefficient hyperparameter. 
As the one-hot labels from the prototypical classifier in the early iterations 
typically would be more noisy, 
the moving average update in Eq.(\ref{qupdate}) 
could gradually discount the previous predictions with a discount factor $\beta$. 
Different from the work \cite{wang2022pico}, we do not set the initial soft-label vectors as uniform vectors
as that will introduce more noise into the soft-labels---%
all labels except one for each instance would be wrong labels.
This makes the soft-label vector not a probability vector as all the elements do not add up to 1. 
It is easy to show that 
the L1-norm of ${\bf q}_i^t$ will always be smaller than 1 
within finite iterations, while approaching 1 when $t\rightarrow \infty$: 
\begin{equation}
    |{\bf q}_i^t|_1 = \beta |{\bf q}_i^{t-1}|_1 + (1-\beta) |{\bf \tilde{y}}_i^t|=(1-\beta)(1+\beta+\beta^2+\cdots+\beta^{t-1})=1-\beta^t.
\end{equation}

The soft-label vectors for the instances in the unlabeled subset $\mathcal{U}$ 
are then further used to fine-tune the target model $f$ by minimizing the following 
{\em soft cross-entropy loss} in the $t$-th iteration:
\begin{equation}\label{loss:lu}
    \mathcal{L}_{u} = \mathbbm{E}_{{\bf x}_i \in \mathcal{U}} 
	\left[ \sum_y - ({\bf q}_i^t)_y \log p(y|{\bf x}_i;{\bf \theta}_f) \right]
\end{equation}
where $({\bf q}_i^t)_y$ denotes the $y$-th element of the vector ${\bf q}_i^t$ 
and it indicates the weight of the cross-entropy loss computed using label $y$. 
If the prototypical classifier keeps assigning the instance ${\bf x}_i$ to the $y$-th class
across iterations, $({\bf q}_i^t)_y$ will 
gradually increase 
to enforce the importance of the pseudo-labeled pair $({\bf x}_i, y)$ for model fine-tuning.
Meanwhile, using soft-label vectors allows the system to consider different label assignment options 
in a weighted manner, which can 
alleviate dramatic oscillations from hard label assignments 
and produce stable model fine-tuning.

By taking both subsets $\mathcal{L}$ and $\mathcal{U}$
into consideration, 
the overall loss minimization 
for the proposed semi-supervised fine-tuning method is shown as follows,
\begin{equation}
    \min_{{\bf \theta}_f} \mathcal{L}_{u} + \lambda \mathcal{L}_{l},
\end{equation}
where $\lambda$ is the trade-off parameter. 
Here we associate the trade-off parameter with the labeled cross-entropy loss 
instead of the soft-labeled cross-entropy loss to avoid 
interactive
influence between parameters $\beta$ and $\lambda$.

\section{Experiments}
\label{sec:exper}

In this section, we conduct experiments to evaluate the proposed source-free inductive domain adaptation method and perform ablation study to 
investigate the separate
contributions from the two components of the proposed method.
As most exiting datasets for domain adaptation do not split training and testing sets, 
we select two large domain adaptation datasets for experiments 
such that the training and testing sets will both have sufficient data.

\subsection{Experimental Setting}
\paragraph{Datasets.} We evaluate our proposed method on two large domain adaptation datasets, \textit{DomainNet} \cite{peng2019moment} and \textit{VisDA2017} \cite{peng2017visda}. 
\textit{DomainNet} is a large-scale classification dataset for domain adaptation. 
It has already split the data in each domain into training and testing sets. 
The original \textit{DomainNet} dataset has 
about 0.6 millions images distributed among 345 categories from 6 domains. 
Here we choose 4 domains with decent oracle performance to use. 
The chosen domains are clipart (c), painting (p), real (r) and sketch (s). 
There are totally 12 domain adaptation tasks among them. 
\textit{VisDA2017} is a large-scale synthetic to real domain adaptive classification dataset. 
There are two domains, a synthetic domain with 152,397 simulation images and a real domain with 55,388 real-word images. 
Both domain contain images from 12 categories and 
we conduct experiments on the synthetic to real domain adaptation task. 
As the original \textit{VisDA2017} dataset does not split training and testing sets, 
we split each domain into training and testing sets with a splitting ratio of $8:2$. To preserve the original class distribution, we split the data from each class 
separately and randomly according to the same ratio. 
We name this split \textit{VisDA2017} dataset as \textit{VisDA2017Split}.

\paragraph{Implementation details.} 
For fair comparisons, 
we follow the previous domain adaptation methods 
and employ the pre-trained ResNet-101 \cite{he2016deep} as the backbone module for both \textit{DomainNet} and \textit{VisDA2017Split} datasets. 
For the same reason,
we replace the final linear layer of the ResNet-101 model with a bottleneck module and a linear layer. The bottleneck module is composed of one linear layer and a 1-dimensional batch normalization layer followed by a ReLU layer. It is used to reduce the feature dimension from 2048 to 256. For all model training and fine-tuning, we use SGD optimizer 
with momentum set as 0.9 and weight decay set as $10^{-3}$. 
The learning rate is scheduled as a cosine decaying function $\eta_i=\eta_1 + 0.5 (\eta_0 - \eta_1) (1 + \cos{(t\pi/N)})$, where $\eta_0, \eta_1, t, N$ are the initial learning rate, stopping learning rate, iteration step and total number of iterations respectively. 
This decaying function decreases the learning rate from $\eta_0$ to $\eta_1$ slowly at the starting and stopping iterations, and rapidly at the intermediate iterations. We set $\eta_0=10^{-2}$ and $\eta_1=10^{-3}$ for the parameters of bottleneck and linear layers. As the backbone is pre-trained on ImageNet, we set $\eta_0=10^{-3}$ and $\eta_1=10^{-4}$ for the parameters of the backbone module.
For source model pre-training, we train the model for 20 epochs 
on the training set and use the testing set to find the best source model by early-stopping. 
For the target model fine-tuning, we set the prediction threshold $p_{th}=0.9$ for both datasets to select high confident pseudo-labels. For soft-label updating, we set two coefficient parameters as $\alpha=0.9$ and $\beta=0.9$. 
In the overall objective, the trade-off parameter $\lambda$ is set to 1 for \textit{DomainNet} dataset and $10^{-2}$ for \textit{VisDA2017Split} dataset.
Our code is available online\footnote{https://github.com/cnyanhao/dmapl}.

\subsection{Results of Source-Free Inductive Domain Adaptation}

\begin{table}[t]
\caption{Test accuracy (\%) on DomainNet dataset (ResNet-101). SF means source-free.}
\label{table:domainnet}
\resizebox{\textwidth}{!}{
\begin{tabular}{l|c|cccccccccccc|c}
\hline
Methods & SF   & c$\shortrightarrow$p & c$\shortrightarrow$r & c$\shortrightarrow$s & p$\shortrightarrow$c & p$\shortrightarrow$r & p$\shortrightarrow$s & r$\shortrightarrow$c & r$\shortrightarrow$p & r$\shortrightarrow$s & s$\shortrightarrow$c & s$\shortrightarrow$p & s$\shortrightarrow$r & Avg. \\
\hline
ResNet-101 \cite{he2016deep} & - & 37.9 & 53.4 & 44.2 & 44.1 & 57.0 & 38.6 & 50.9 & 48.8 & 37.7 & 52.8 & 37.3 & 47.6 & 45.9 \\
AdaMatch \cite{berthelot2022adamatch} & \xmark & 45.3 & 56.0 & 60.2 & 35.3 & 47.6 & 42.9 & 46.5 & 48.1 & 49.1 & 46.5 & 41.0 & 42.4 & 46.7 \\
MCC \cite{jin2020minimum} & \xmark & 37.7 & 55.7 & 42.6 & 45.4 & 59.8 & 39.9 & 54.4 & 53.1 & 37.0 & 58.1 & 46.3 & 56.2 & 48.9 \\
CDAN \cite{mingsheng2018, dalib} & \xmark & 40.4 & 56.8 & 46.1 & 45.1 & 58.4 & 40.5 & 55.6 & 53.6 & 43.0 & 57.2 & 46.4 & 55.7 & 49.9 \\
CDAN+SDAT \cite{rangwani2022closer} & \xmark & 41.5 & 57.5 & 47.2 & 47.5 & 58.0 & 41.8 & 56.7 & 53.6 & 43.9 & 58.7 & 48.1 & 57.1 & 51.0 \\
\hline
SHOT \cite{liang2020we} & \cmark & 45.6 & 63.4 & 49.1 & 35.1 & 64.1 & 21.0 & 57.1 & 51.1 & 44.0 & 61.2 & 47.6 & 62.0 & 48.4 \\
SSFT-SSD \cite{yan2021source} & \cmark & 41.9 & 57.5 & 46.5 & 47.6 & 59.6 & 42.6 & 55.4 & 51.9 & 42.0 & 58.4 & 45.2 & 55.7 & 50.4 \\
DMAPL (Ours) & \cmark & {\bf 46.0} & {\bf 63.7} & {\bf 49.1} & {\bf 53.2} & {\bf 64.2} & {\bf 46.0} & {\bf 61.6} & {\bf 55.4} & {\bf 47.8} & {\bf 64.1} & {\bf 50.3} & {\bf 63.5} & {\bf 55.4} \\
\hline
Oracle & - & 71.1 & 83.4 & 70.0 & 78.4 & 83.4 & 70.0 & 78.4 & 71.1 & 70.0 & 78.4 & 71.1 & 83.4 & 75.7 \\
\hline
\end{tabular}
}

\bigskip

\caption{Test accuracy (\%) on VisDA2017Split dataset (ResNet-101). SF means source-free.}
\label{table:visdasplit}
\resizebox{\textwidth}{!}{
\begin{tabular}{l|c|cccccccccccc|cc}
\hline
Methods & SF & plane & bcycl & bus & car & horse & knife & mcycl & person & plant & sktbrd & train & truck & Macro & Micro \\
\hline
ResNet-101 \cite{he2016deep} & - & 76.7 & 23.9 & 48.1 & 68.0 & 67.8 & 6.5 & 86.0 & 20.6 & 71.8 & 23.9 & 85.0 & 8.4 & 48.9 & 54.1 \\
CDAN \cite{mingsheng2018, dalib} & \xmark & 92.7 & 73.5 & 80.0 & 46.4 & 90.2 & 93.2 & 86.1 & 78.4 & 83.8 & 87.3 & 83.2 & 38.3 & 77.8 & 73.7 \\
MCC \cite{jin2020minimum} & \xmark & 92.2 & 79.4 & 79.0 & 71.7 & 92.1 & 93.0 & 89.9 & 79.0 & 88.2 & 91.0 & 82.1 & 50.8 & 82.4 & 80.0 \\
\hline
SHOT \cite{liang2020we} & \cmark & 77.7 & {\bf 85.8} & 80.2 & 54.2 & 90.2 & 63.4 & {\bf 82.1} & 73.5 & 88.9 & 80.5 & 83.1 & 54.8 & 76.2 & 73.8 \\
SSFT-SSD \cite{yan2021source} & \cmark & 94.5 & 84.9 & {\bf 80.9} & 49.9 & 91.2 & 66.8 & 77.0 & 75.4 & 81.3 & 86.2 & {\bf 89.4} & 50.4 & 77.3 & 73.6 \\
DMAPL (Ours) & \cmark & {\bf 95.6} & 84.5 & 78.9 & {\bf 58.7} & {\bf 92.4} & {\bf 96.6} & 80.8 & {\bf 82.5} & {\bf 90.3} & {\bf 88.6} & 87.8 & {\bf 59.1} & {\bf 83.0} & {\bf 79.1} \\
\hline
Oracle & - & 98.2 & 94.7 & 89.5 & 88.0 & 98.7 & 96.4 & 93.6 & 92.8 & 98.0 & 96.5 & 93.4 & 72.6 & 92.7 & 91.5 \\
\hline
\end{tabular}
}
\end{table}

To evaluate the proposed Dual Moving Average Pseudo-Labeling (DMAPL) method, we conduct source-free inductive domain adaptation experiments on the two datasets, \textit{DomainNet} and \textit{VisDA2017Split}. 
The results are reported in Table~\ref{table:domainnet} and Table~\ref{table:visdasplit}. With the pre-trained source model, we first evaluate it on the unlabeled target testing set and 
produce the source-only result on each task, 
which demonstrates
the performance of the source model without any adaptation process and 
is used as a comparison baseline.
We also report the target-supervised results as an {\em oracle} reference, 
where the model is trained on the labeled target training set and evaluated on the target testing set. 
To compare with the existing domain adaptation methods, 
when there are existing inductive results, 
we cite the results directly; 
e.g., the test results on the \textit{DomainNet} dataset for CDAN \cite{mingsheng2018, dalib}, MCC \cite{jin2020minimum}, SDAT \cite{rangwani2022closer} and AdaMatch \cite{berthelot2022adamatch}. 
Otherwise, we run the methods in the inductive setting to produce fair comparisons. 
As there are no reported SFDA results in the inductive setting, we selected two SFDA methods with online code,
SHOT and SSFT-SSD, to compare with our proposed method in the same inductive setting.
We produce the comparison results on the \textit{VisDA2017Split} dataset in a similar way.

Table~\ref{table:domainnet} reports the domain adaptation test accuracy results
 for the 12 domain adaptation tasks and the task average results on the \textit{DomainNet} dataset. 
This table comprises two parts of results. The upper part includes the {\em source-only} results denoted as 
{\em ResNet-101} and the results of the several UDA methods that use both the labeled source data
and unlabeled target data. 
The lower part reports the results of our proposed method (DMAPL)
and two existing source-free domain adaptation methods---the results are produced in the inductive SFDA setting. 
The bottom line reports the target supervised results denoted as {\em oracle}.
First, we can see
our proposed method improves the testing accuracy over source-only by a large margin on every task, and on average a 9.5 percentage points (pp) gain can be observed. 
This is a remarkable 
improvement on this challenging dataset.
Second, comparing with the several unsupervised domain adaptation methods, our method outperforms 
all of them even though the source-free setting is more restrictive than the UDA setting. 
This shows the effectiveness of the proposed semi-supervised fine-tuning framework 
on exploiting the unlabeled target training data. 
Similar results have been widely observed in the previous source-free domain adaptation works as well \cite{liang2020we, lee2022confidence}.
Third, our method also outperforms both of the two SFDA comparison methods
by large margins. Our method achieves the state-of-the-art performance on all the 12 tasks. 
Comparing with the oracle results, we can see 
there is still a large space for further improvement.
In addition, there are also some other interesting observations. 
For example, the two UDA methods,
MCC and AdaMatch, have been previously shown to perform better than CDAN in the transductive domain adaptation setting \cite{jin2020minimum, berthelot2022adamatch}, while here CDAN achieves better results 
than MCC and AdaMatch
in the inductive setting.
The two source-free domain adaptation comparison methods, 
SHOT and SSFT-SSD, have been shown to outperform the source-only baseline by large margins 
in the transductive setting in the literature, while here they can only achieve small performance gains
in the inductive setting. 
These observations might
inspire future works to 
shift the focus of domain adaptation from the transductive setting to the inductive setting.

Table~\ref{table:visdasplit} reports the test accuracy results on the 12 categories, 
the macro accuracy and the micro accuracy 
for the synthetic-to-real domain adaptation task 
on the \textit{VisDA2017Split} dataset. 
Here the macro accuracy is the test accuracy averaged over the 12 classes and 
the micro accuracy is the test accuracy averaged over all test instances. 
First, comparing with the source-only baseline, 
our proposed method improves the macro accuracy by 34.1 pp and improves the micro accuracy by 25 pp. 
Second, our proposed method also outperforms all the UDA methods, though UDA has the source data available. 
Finally, our method outperforms the two SFDA comparison methods by large margins 
in terms of both macro accuracy
and micro accuracy
under the inductive setting. 
Overall, the proposed method achieves state-of-the-art performance on 8 out of the 12 categories, 
as well as in terms of macro accuracy and micro accuracy. 
But still there is large improvement space for future works by comparing with the oracle results. 

\subsection{Ablation Study and Hyper-Parameters Analysis}

\begin{table}[t]
\centering
\caption{Ablation study: Average accuracy ($\%$)}
\label{tab:ablation}
\resizebox{0.9\textwidth}{!}{
\begin{tabular}{l|cccc}
\hline
Ablation   & Split target & DomainNet & VisDA2017 (Micro) & VisDA2017 (Macro) \\
\hline
Source-only & \xmark & 45.9 & 54.1 & 48.9 \\
Naive pseudo-labeling \cite{lee2013pseudo} & \xmark & 47.8 & 61.2 & 55.7 \\
Soft-label updating & \xmark & 52.4 & 78.3 & 82.2 \\
DMAPL & \cmark & 55.3 & 79.1 & 83.0 \\
\hline
\end{tabular}
}
\end{table}


\paragraph{Ablation study.} 
We investigate the contributions 
of the two components of the proposed method: the target training data split
and the dual moving average based soft pseudo-labeling. 
We compare the proposed method, DMAPL, with the Source-only baseline and two variants:
{\em naive pseudo-labeling} and {\em soft-label updating}. 
{\em Naive pseudo-labeling} does not split the target training data, but
assigns pseudo-labels to the unlabeled target training data at the beginning of each epoch 
and uses them for target model fine-tuning;
{\em Soft-label updating} drops the target training data splitting step from the proposed method
and uses all target training data for soft pseudo-label based fine-tuning.  
As shown in Table~\ref{tab:ablation}, 
comparing with the complete method DMAPL, 
{\em soft-labeling updating} without target data split degrades the performance on both datasets.
This verifies the usage of the selected target training instances as fixed labeled data
during model fine-tuning.  
Meanwhile, the proposed method DMAPL outperforms {\em naive pseudo-labeling} by large margins on both datasets. 
This shows the effectiveness of the proposed soft-label updating method. 

\begin{figure}[t]
\centering
\begin{subfigure}
  \centering
  \includegraphics[width=0.3\textwidth]{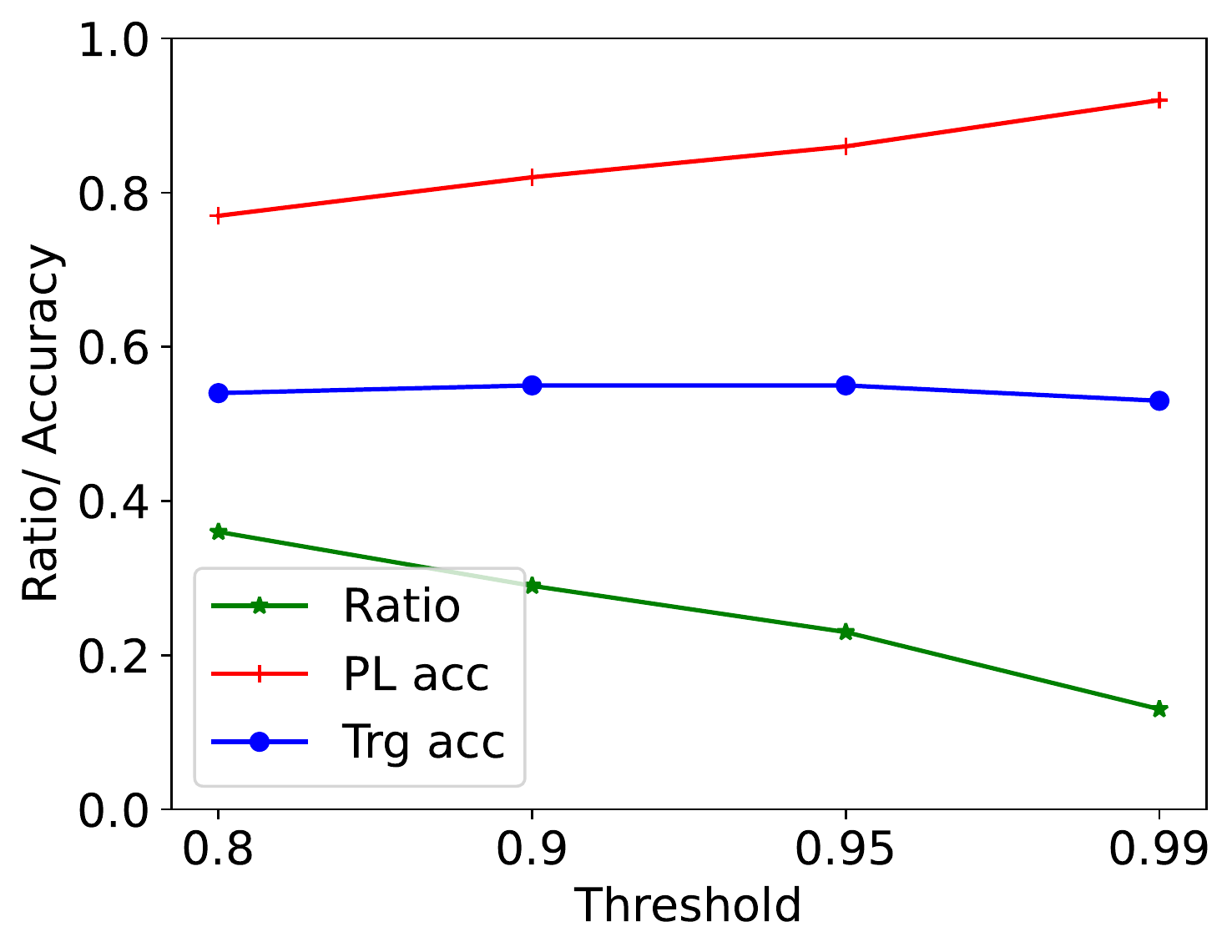}
\end{subfigure}
\hfill
\begin{subfigure}
  \centering
  \includegraphics[width=0.3\textwidth]{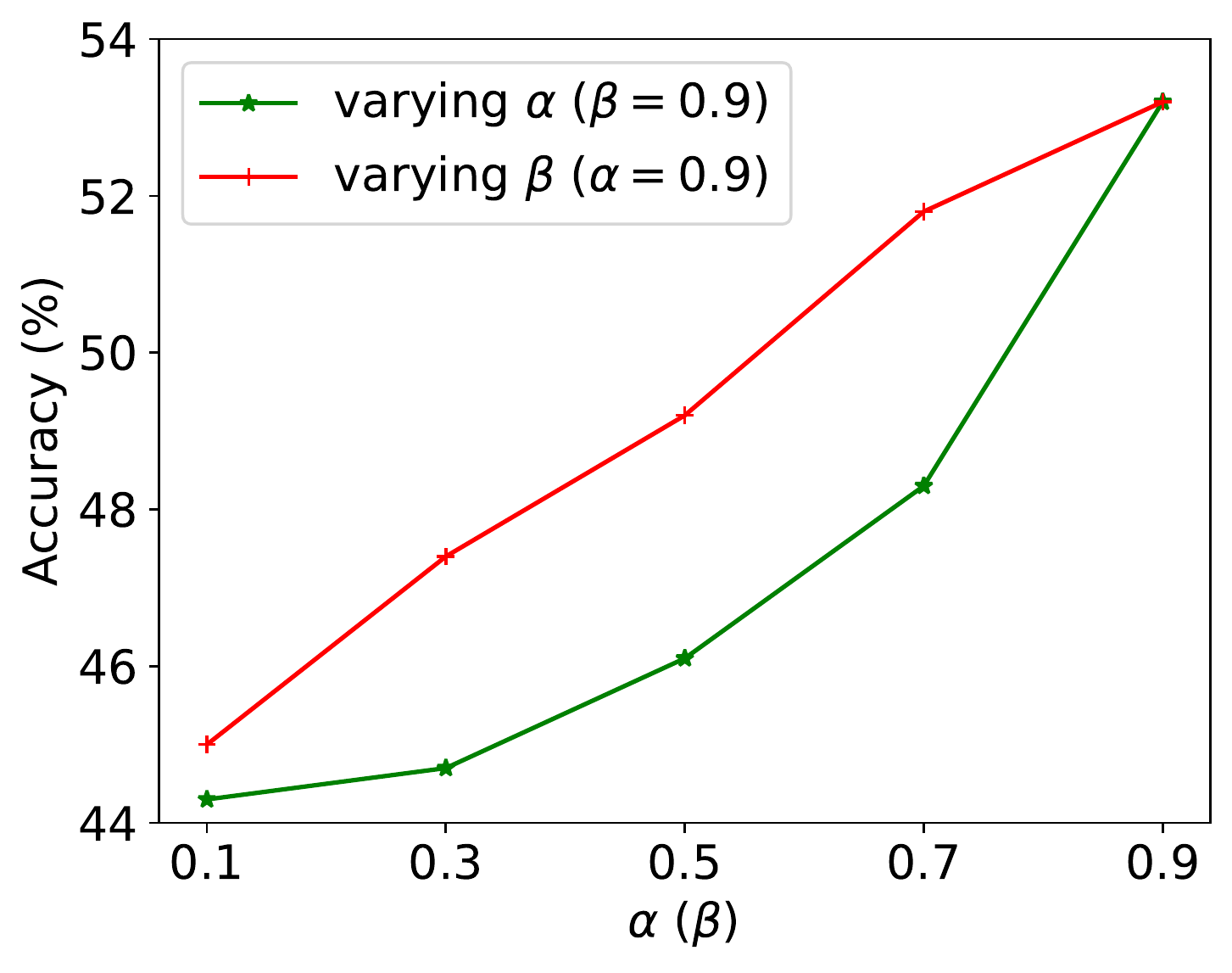}
\end{subfigure}
\hfill
\begin{subfigure}
  \centering
  \includegraphics[width=0.3\textwidth]{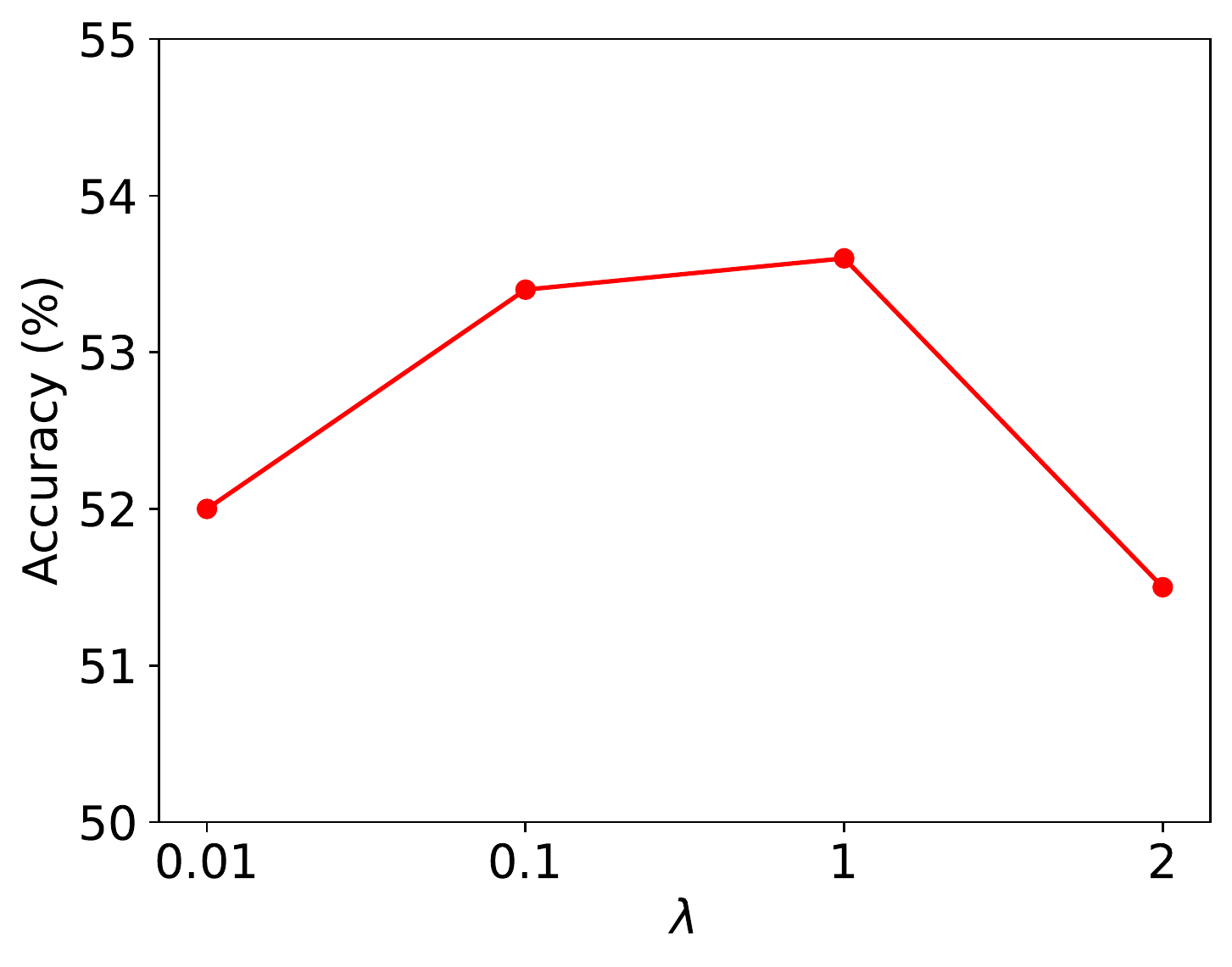}
\end{subfigure}
	\caption{Hyper-parameters analysis. Experiments are conducted on task p $\to$ c of DomainNet dataset. Left: Confident subset splitting {\em Ratio}, accuracy of the pseudo-labels for the confident subset ({\em PL acc}) and test set accuracy ({\em Trg acc}) for different thresholds ($p_{th}$). Middle: Test set accuracy for different choices of coefficient parameters $\alpha$ and $\beta$. Right: Test set accuracy for different choices of the trade-off parameter $\lambda$.}
\label{fig:hyperparam}
\end{figure}

\paragraph{Confidence threshold.} 
The left figure of Figure~\ref{fig:hyperparam} shows the confident subset splitting {\em Ratio}
($|\mathcal{L}|$/|$\mathcal{X}_T$|), accuracy of the pseudo-labels for the confident subset $\mathcal{L}$
({\em PL accu}),
and test set accuracy ({\em Trg acc}) values for different threshold ($p_{th}$) choices on the task 
p $\to$ c of \textit{DomainNet}. 
We can observe that higher threshold makes the pseudo-labels for the high confident subset cleaner
(higher {\em PL acc})
at the expense of less instances in the high confident subset (lower Ratio). 
However, similar testing accuracy values ({\em Trg acc}) can be observed with different threshold choices. The reason might be credited to the effectiveness of the proposed soft-label updating approach. This indicates that our proposed method is not very sensitive to the $p_{th}$ value as long as it is reasonably large ($0.8\leq p_{th}<1$).

\paragraph{Coefficient parameters.} 
The middle figure of Figure~\ref{fig:hyperparam} illustrates the test accuracy values against different choices of the coefficient parameters $\alpha$ and $\beta$ on the task p $\to$ c of \textit{DomainNet},  
where $\alpha$ and $\beta$ control the updating degrees for centroid update and 
soft-label update respectively---lower values indicate larger updating degrees.  
It is obviously that higher values 
achieve better performance for both variables. 
This indicates that slower updates of the centroid vectors and soft-labels are more beneficial for the proposed method, due to better training stability. 
This coincides with results from other moving average 
cases \cite{tarvainen2017mean} as well. 

\paragraph{Trade-off parameter.} 
The right figure of Figure~\ref{fig:hyperparam} demonstrates the test set accuracy values for different choices of the trade-off parameter $\lambda$ on the task p $\to$ c of \textit{DomainNet}.
We can observe that the proposed method is not very sensitive to this trade-off parameter
within a reasonable range of values $[0.1, 1]$, 
while the optimal $\lambda$ value is 1. 

\section{Conclusion}
\label{sec:concl}
This paper addresses a newly proposed source-free inductive domain adaptation problem. 
We proposed a new semi-supervised fine-tuning method based on dual moving average pseudo-labeling. 
To prevent over adaptation of the source model to the unlabeled target training data,
we proposed to split the unlabeled target training data into 
a pseudo-labeled subset with high prediction confidence and an unlabeled subset with less confident predictions
based on the pre-trained source model. 
The model is fine-tuned in a semi-supervised manner by using
the pseudo-labeled subset as trusty supervision, and using
the unlabeled subset with soft-labels produced by dual moving average pseudo-labeling.
The experimental results show that our method achieves state-of-the-art performance and outperforms previous methods by large margins for source-free inductive domain adaptation. 
The experiments also indicate that some previous domain adaptation methods 
that are effective in the transductive setting can achieve less performance gains in the inductive setting. 
This is expected to motivate future works
in the more realistic inductive domain adaptation settings.

\bibliography{mybib}
\end{document}